# Deep Learning for Multi-Level Detection and Localization of Myocardial Scars Based on Regional Strain Validated on Virtual Patients

Müjde Akdeniz[1,2], Claudia Alessandra Manetti[3], Tijmen Koopsen[3], Hani Nozari Mirar[1,2], Sten Roar Snare[1], Svein Arne Aase[1], Joost Lumens[3], Jurica Šprem[1], and Kristin Sarah McLeod[1]

[1]GE Vingmed Ultrasound, Oslo, Norway
[2]Department of Informatics, University of Oslo, Oslo, Norway
[3]Cardiovascular Research Institute Maastricht (CARIM) at Maastricht University Medical Center, Maastricht, The Netherlands

Corresponding author: Müjde Akdeniz (e-mail: mujde.akdeniz@ge.com).

This project has received funding from the European Union's Horizon 2020 research and innovation programme under the Marie-Sklodowska-Curie grant agreement No 860745.

**ABSTRACT** How well the heart is functioning can be quantified through measurements of myocardial deformation via echocardiography. Clinical assessment of cardiac function is generally focused on global indices of relative shortening, however, territorial, and segmental strain indices have shown to be abnormal in regions of myocardial disease, such as scar. In this work, we propose a single framework to predict myocardial disease substrates at global, territorial, and segmental levels using regional myocardial strain traces as input to a convolutional neural network (CNN)-based classification algorithm. An anatomically meaningful representation of the input data from the clinically standard bullseye representation to a multi-channel 2D image is proposed, to formulate the task as an image classification problem, thus enabling the use of state-of-the-art neural network configurations. A Fully Convolutional Network (FCN) is trained to detect and localize myocardial scar from regional left ventricular (LV) strain patterns. Simulated regional strain data from a controlled dataset of virtual patients with varying degrees and locations of myocardial scar is used for training and validation. The proposed method successfully detects and localizes the scars on 98% of the 5490 left ventricle (LV) segments of the 305 patients in the test set using strain traces only. Due to the sparse existence of scar, only 10% of the LV segments in the virtual patient cohort have scar. Taking the imbalance into account, the class balanced accuracy is calculated as 95%. The performance is reported on global, territorial, and segmental levels. The proposed method proves successful on the strain traces of the virtual cohort and offers the potential to solve the regional myocardial scar detection problem on the strain traces of the real patient cohorts.

**INDEX TERMS** Deep learning, fully convolutional network, myocardial scar, strain

## I. INTRODUCTION

Echocardiography uses ultrasound technology to visualize different structures of the heart, offering users various methods to quantify its functioning. One of the main functions of the heart is to pump blood throughout the body. One way to evaluate the pumping abilities of the heart is to quantify the changes in shapes i.e., deformation of its chambers. 2D Speckle Tracking Echocardiography (STE) is a technique for tracking the speckles in the myocardial tissue of the heart chambers throughout the heart cycle. Strain and strain rate can be measured using this tracking technology. Strain measurements enable the assessment of the systolic and diastolic functions of the LV. Longitudinal strain is calculated at each time point of the cardiac cycle as the fractional length change of a myocardial segment relative to its length at end diastole, as measured via 2D STE.

Apical four chamber (4CH), apical two chamber (2CH) and apical long axis (APLAX) views are necessary for calculating the longitudinal strain for 18 LV segments. Each view yields six segmental strain traces, resulting in 18 traces in total. Figure 1 shows an example of a strain measurement result. Since strain is the measure of relative length change, in a complete cycle, the measurements should start and end at zero.

Bull's eye in the figure shows the peak systolic strain for 18 LV segments [1].

Segmental strain traces contain diagnostic information which can aid the clinician in therapeutic decision making. An important example is the use of strain imaging for detecting myocardial scar in heart failure (HF) patients considered for cardiac resynchronization therapy (CRT). Previous studies have demonstrated that presence of myocardial scar can cause non-response to CRT [2] or even induce life-threatening arrhythmias upon pacing [3],[4]. It has additionally been shown that posterolateral scar is associated with less favorable outcome after CRT [5], [6]. Not only the size of scar but also its location should therefore be taken into account when implanting a CRT device.

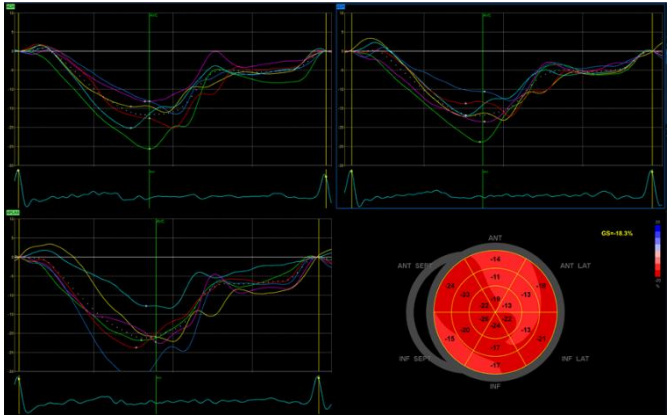

*Figure 1 Example of a strain measurement shown in EchoPAC (GE Healthcare, Horten, NO), demonstrating the deformation of the LV throughout the heart cycle. Strain traces are measured from 4CH (top left), 2CH (top right) and APLAX (bottom left) views.*

In current clinical practice, myocardial scar can readily be detected by late gadolinium enhancement (LGE) cardiac magnetic resonance (CMR) imaging. The use of LGE-CMR, however, has several disadvantages, including that it is relatively costly, it involves administration of a contrast agent, and it is unsuitable for patients with claustrophobia or previous device implantation. Using non-invasive echocardiographic strain imaging could provide a faster, cheaper, and more widely applicable alternative for myocardial scar detection.

Clinical studies on strain analysis point to the difference between the strain traces of a scarred and healthy region in heart. D'Andrea et al. analyzed 2D STE strain values from 45 patients with ischemic dilated cardiomyopathy (DCM) at segmental and global level for understanding effects of strain and scar on CRT response [7]. Peak systolic strain values were significantly different for infarcted and non-infarcted segments. Global longitudinal strain (GLS) for each subject was calculated as the average of peak strain for all segments and were reduced for the patients with DCM when compared to healthy subjects. Having access to CMR images to assess the location and size of the scars, they found a correlation between reduced strain and transmural scar extent.

Candidates for CRT typically have a left bundle branch block (LBBB) contraction pattern [8], which complicates the use of peak strain for myocardial scar detection. In the early activated septum, peak strain can be reduced while contractile strength is unaffected, while the late activated lateral wall can have increased peak strain due to its Frank-Starling behavior. Consequently, in LBBB, quantification of peak strain does not accurately reflect myocardial contractile properties, and there is a need for a better approach to detect scar which does not rely on a single strain feature. There is currently a lack of tools to automate the assessment of the strain data to cope the task of identifying myocardial regions with scar. In this study, we therefore aim to develop a clinical tool which automatically identifies regions with scar from a patient's set of complex myocardial strain patterns.

The relationship between myocardial scar and regional longitudinal strain was validated in [9] on 96 patients with coronary artery disease. Strain traces were measured with feature tracking algorithm of the LGE CMR images. They observed a reduced magnitude of regional longitudinal strain in scarred areas when compared to non-scarred areas.

With the growing size of clinical patient data, there is a growing clinical need for diagnostic tools that could leverage the advances in the field of machine learning. A machine learning model is developed to automatically identify hypertrophic cardiomyopathy (HCM) patients without myocardial fibrosis [10]. Although the authors used functional and morphological features of myocardium and not strain traces, they point to the importance of finding patterns in strain traces of patients with non-ischemic fibrosis, such as HCM patients. They suggest that their machine learning algorithm could benefit from adding the strain parameters to the input.

There are also applications of machine learning methods that target the automatic interpretation of strain traces for better diagnosis. Tabassian and coauthors used Principal Component Analysis (PCA) to extract temporal features from the strain traces from acute myocardial infarction (MI) patients; patients with suspected ischemic heart disease without any scar and healthy subjects [11]. They concatenated each temporal feature set of 18 LV segments to obtain a single spatio-temporal feature set per patient. Two independent K Nearest Neighbor (KNN) classifiers were built for strain and strain rate data separately to distinguish healthy and MI subjects. The proposed machine learning framework yielded 87% accuracy, which was significantly higher compared to overall 70 % accuracy of the expert readers.

Loncaric et al. showed that unsupervised methods can help clinicians understand phenotypes purely by automatic interpretation of echocardiography data, eliminating the possible bias towards any diagnosis or outcome [12]. They used Multiple Kernel Learning of velocity traces from Pulsed Wave Doppler in addition to strain traces from 2D STE, to represent a hypertensive patient cohort in a lower dimensional space. Using regression methods, they estimated average strain and velocity profiles and were able to detect subtle differences between LV basal septal strain profiles from opposing regions of the hypertensive patient space. They divided the hypertensive space into four groups using k-means clustering, where the first two groups correlated with healthy and transitional hearts, while the third group had clear remodeling due to high blood pressure, and the final group correlated with female patients with other co-morbidities.

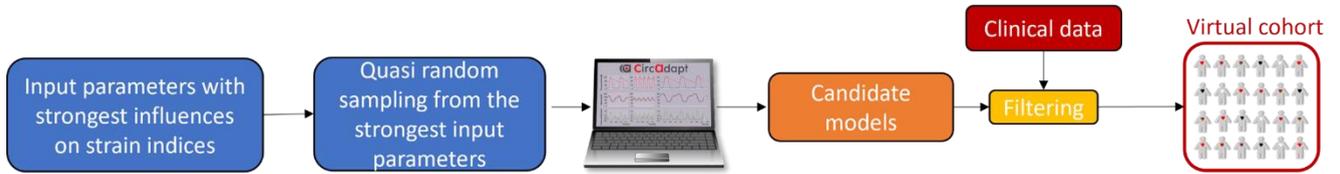

*Figure 2 The virtual patient simulation pipeline with the CircAdapt model.*

Similarly, Cikes at al. uses strain traces along with volume traces, temporal deformation vector and clinical parameters to predict response to CRT on a large set of heart failure (HF) patients [13]. The authors were able to divide the patients into four phenogroups using unsupervised machine learning methods and identify the responder and non-responder patients.

In this work, our aim is to automate the analysis of the segmental strain traces, to recognize subtle changes and patterns that suggest scar infiltration, including those that may be unnoticeable to the examiner beside other clinical tasks in their heavy workload. We propose a deep learning model and train it on a set of virtual patient cohort as proof of concept for this task. Using a virtual cohort of synthetic patients to test the proposed approach enables both access to an automatically labeled strain dataset with known disease substrates and a controlled dataset where labels are certain. We use the proposed approach to identify the patients with scarred segments and quantify the scar. We propose that using anatomical representation of temporal strain patterns instead of peak systolic segmental strain and/or GLS brings additional value when assessing the extent of the scar in LV.

We are investigating whether we can identify the presence of myocardial scar at a global, territorial, and segmental level based on regional strain alone.

There are three main contributions of the study:
1. Using a bull's eye representation of the strain traces as an input to the model by converting a 2D signal to an image space that captures the inter-connectivity between regions.
2. Using a neural network model to classify the strain traces to detect scar at a global, territorial, and segmental level.
3. Leveraging synthetically generated regional strain data to test the methods under well-controlled conditions.

While there are numerous studies using deep learning technologies on medical data; to the best of our knowledge, this is the first study to use neural networks to classify strain data from virtual patients to detect scar at multiple levels.

**II. MATERIALS AND METHODS**

A virtual patient cohort is created to generate realistic synthetic strain data, which gives the advantage of having the ground truth labels on presence, location, and extension of scar available. Having the labels available for relatively big amount of data compared to what is usually available in clinical practice enables the use of a supervised deep learning approach. The bull's eye representation of strain data is used for preserving the spatial information from the regional strain traces and creates a template for representing strain data as a suitable input to CNN architectures. Finally, a fully convolutional neural network (FCN) is employed to predict the scar existence at segmental, territorial, and global levels.

*A. CIRCADAPT VIRTUAL PATIENT DATASET*

The CircAdapt computational model of the human heart and circulation [14], [15] developed at Maastricht University, was used to simulate regional myocardial mechanics and global hemodynamics in an 18-segments model of the LV. To generate the virtual population of heart failure patients, 7000 parameters set were generated using the Sobol-Low Discrepancy sequence [16]. Each parameter set contained different model parameters describing global cardiac pump function and regional systolic and diastolic tissue properties. These parameters were chosen because they are the most sensitive in determining ventricular strain patterns in patients with electromechanical dyssynchrony and myocardial infarction. The methodology to simulate virtual patients with myocardial infarction using a lumped two-compartment modeling approach integrated into the CircAdapt model is previously introduced in [17]. The resulting cohort was filtered with global hemodynamic parameter values and strain indices representative of a CRT cohort, found in literature, guidelines [18], and clinical data. The final cohort consisted of 3043 virtual patients affected by left bundle branch block (LBBB) and possible myocardial infarction (MI) in one of the three coronary artery territories.

For the experiments in this work, significant input parameters of the CircAdapt model are left ventricle and right ventricle global contractility (Pa), delay of mechanical activation (sec) and relative volume of myocardial scar (%). The output of interest from the model is the set of strain traces of the virtual patients.

The advantage of using simulated data is three-fold:
Data availability: Any number of patients can be generated with desired abnormalities.

Label availability: With predefined threshold values that are used to tune the input parameters, we generate the scar labels automatically and objectively. There is no manual process involved in labeling.

Transfer learning: The algorithms that are developed for recognizing patterns in strain traces are not specific for virtual patients and can be re-trained on the strain traces of real patient data with the help of transfer learning techniques.

Each patient in the dataset has different levels of extension of scar depending on the volume fraction of the scarred region. Since predicting the scar extension of the patients is out of the scope of this study, labels are generated by mapping the volume fractions to binary labels to indicate the existence of scar. Using a one hot encoding setup, the LV segments with any non-zero volume fraction of scar are labeled with one and the LV segments with no-scar were labeled with zero. The American Heart Association's (ASE) 18 segment model for the LV is used in this project [1].

### B. TEMPORAL PREPROCESSING OF THE STRAIN DATA

The heart rate values for each patient are different in the virtual patient dataset. Having variability in the duration of the heart cycles requires a preprocessing operation to handle different number of data points present in each trace since input data lengths need to be consistent for the use of CNNs.

In an earlier study [11], Tabassian et al. used purely the LV longitudinal strain and strain rate traces to classify infarct. To overcome the issue of variable number of data points, they interpolated the strain traces to the average number of data points per cardiac phase over all healthy subjects. In the proposed approach, we also use a time series re-sampling method, but instead of resampling to the average number of data points in patient strain traces, we use a reference strain length to resample to. In addition, we use the aortic valve closure (AVC) time point to resample the systole and diastole phases separately before merging them back into one signal.

We empirically set the number of data points to 500 to obtain equal length feature vectors for all patients. Figure 3 demonstrates the differences in strain patterns of two sample patients with and without the scar, after the preprocessing operations are applied on the traces.

### C. BULL'S EYE SPATIAL REPRESENTATION OF STRAIN UNWRAPPED TO AN IMAGE GRID

#### 1) FROM STRAIN TRACES TO BULL'S EYE

We integrate the spatial locations of the LV segments into the input tensor. The spatial locations of the LV segments are represented with the well-known bull's eye representation from clinics.

#### 2) BULL'S EYE CIRCULAR GRID UNWRAPPED TO A RECTANGULAR GRID

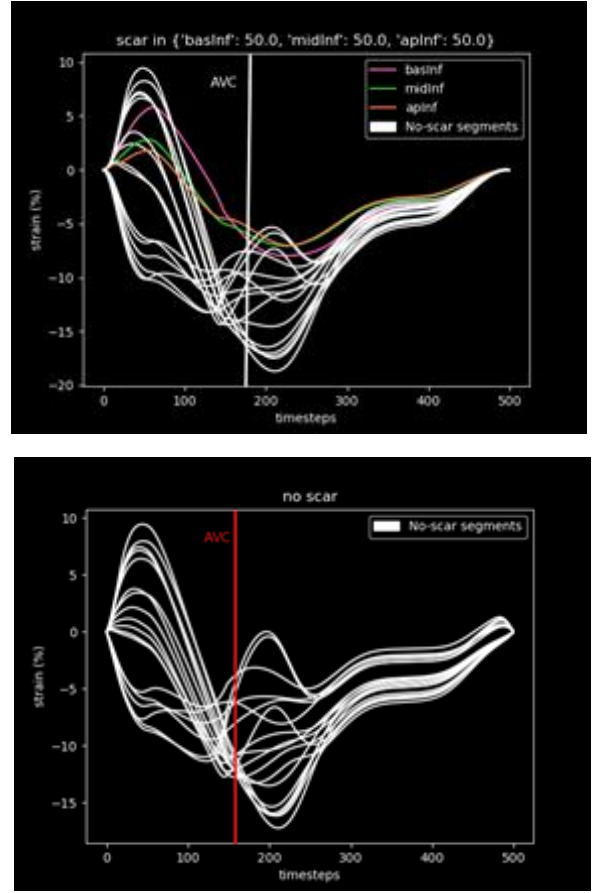

*Figure 3 Comparison of the strain traces of patients with scar (top) and without scar (bottom) after temporal preprocessing of the systolic and diastolic phases separately, to yield an equivalent length of the strain data. Reduced contractility in systole is observed for some of the segments of the patients with scar. As CircAdapt virtual patient dataset contains patients with LBBB, one can notice abnormal behavior in the strain traces of the patient without scar (bottom).*

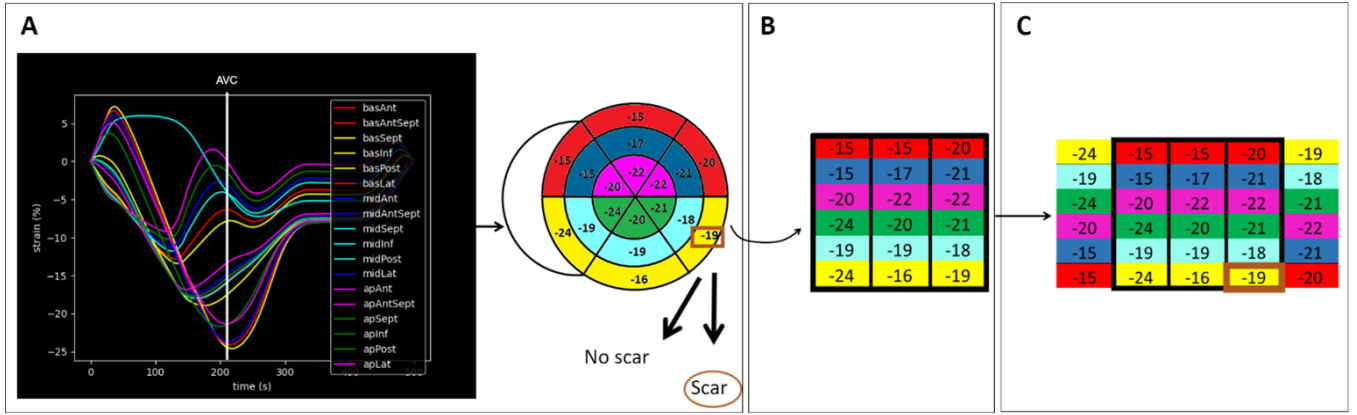

*Figure 4 Spatial representation of the strain. A) Deformation patterns of the left ventricle to a 2D bullseye representation. It should be noted that the bull's eye captures the strain values from a single point in time. B) Unwrap circular bullseye to rectangular 2D matrix. C) Pad the bullseye matrix to imitate the circularity present in the bull's eye shape.*

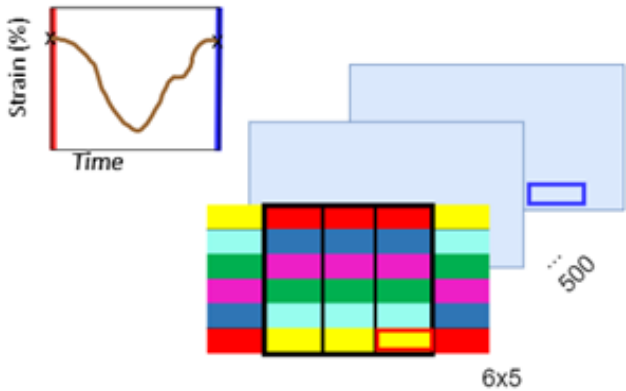

*Figure 5 Temporal representation of the strain. The channels of the input layer in the FCN are used for capturing the time axis of the strain data, eliminating the need for using a 3D CNN architecture. It should be noted that the strain trace shown in this figure is from a single LV segment. Every patient has one trace for each segment throughout the cycle, resulting in 18 traces for the 18-segment model of the LV.*

Creating an image representation of the strain helps us leverage CNNs for strain classification and represent the patient data anatomically. We unwrap the circular bull's eye into a 2D matrix.

3) HORIZONTAL PADDING FOR CIRCULARITY

Padding is applied to preserve the circularity of the real LV geometry. The horizontal padding is used since it reflects the continuity of the neighboring segments.

Strain data holds information along the time axis, and therefore has to be stored in a 3D structure with two spatial dimensions for LV segments (Figure 4) and one temporal dimension for the time datapoints of the heart cycle (Figure 5). Representing the strain traces with bull's eye and unwrapping it to a 2D grid yields a robust template for patient strain data which can be used for training a 2D CNN architecture of choice for any given strain classification task.

**D. STRAIN CLASSIFICATION MODEL**

The proposed method focuses both on identifying patients with scar and on localizing the scar for guiding the clinician to improve treatment plans. An example use case for the former could be to group the patients based on the attention they need due to possible scar in their LV regions i.e. patients with scar may need to be more regularly followed up than those without. An example use case for the latter could be to predict the response to CRT when deciding for CRT candidates, based on the knowledge of which coronary artery territory (CAT) is affected and therefore may not benefit from corresponding the proximity of the pacemaker lead to the scarred region. While the feature extraction and classification layers of any CNN classifier could be used for the former problem, a specific architecture is required to solve the latter problem.

A simplified version of the FCN introduced in [19] is implemented for this task to tackle the scar localization problem, which outputs the prediction in the form of a binary mask of LV segments as shown in Figure 6. The original FCN architecture predicts the pixel-wise scores for each class, which results in an output layer with channel dimension N, where N is the number of classes in their image segmentation task. Similarly, the proposed architecture predicts the segment-wise scores for scar and no-scar classes, which results in an output layer with channel dimension two.

The channels in the input layer represent the time steps from the strain traces. In each channel, strain values from all 18 segments are represented in the 2D tensor for that time step. Since the strain traces carry the information for each LV segment over the entire heart cycle, the data is high dimensional in the time axis. Using 2D CNNS with many input channels to store the temporal data, we eliminate the need of using computationally expensive 3D CNN architectures.

Three layers with 2D convolution operations are followed by a 2D transpose convolution operation as the last layer, to convert the feature maps back to the original shape of the unwrapped bull's eye grid. The output of the scar detection

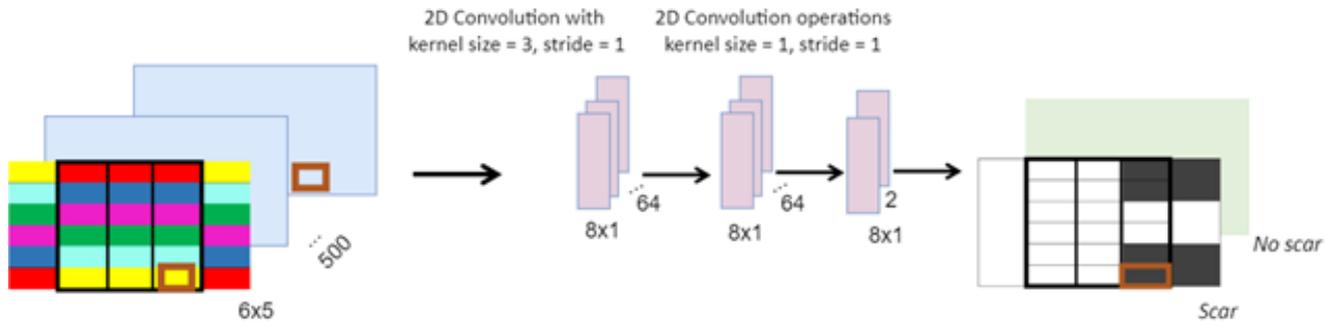

*Figure 6 Local scar detection pipeline for LV segmental strain traces. The temporal representation of the strain is fed to the CNN for convolution operations.*

FCN is a stack of two tensors with the prediction values of each segment having scar (first channel) and no scar (second channel) as shown in Figure 6. The class with the highest prediction score for a given segment is assigned as the predicted class for that segment.

In addition to detection, we quantify the scar extent in terms of number of affected LV segments and their locations. After obtaining the scar predictions at the segmental level, we post-process the results to generate predictions for CAT and patient levels. A CAT is considered to have scar if any of its corresponding segments has scar; similarly, a patient is considered to have scar if any of their LV segments has scar.

## III. RESULTS

The original dataset of 3043 virtual patients is partitioned into development and test sets with ratios of 90% and 10% respectively, taking the number of scarred segments into account for stratification. 80% and 20% of the development set was used for training and validation respectively. After setting the padding configuration to no-padding, training and validation sets are scaled by sampling 50%, 75% and 100% of the patients to assess the effect of dataset size on scar prediction performance. The validation loss is minimum when the model is trained on 100% of the dataset. The models trained on 50%, 75% and 100% of the dataset yield balanced accuracy scores of 92%, 93% and 95% respectively.

In addition, to assess the effect of padding applied on 2D bulls eye template, the scale of the dataset is set to constant percentage of 50% and the model is trained with bull's eye templates with no padding and horizontal padding configurations. The minimum validation loss is obtained from the horizontal padding. After observing the experiment setups with the minimum validation losses, the size of the dataset to be used and the padding configuration of the bull's eye template is determined accordingly. The experiment is then carried out on the entire dataset with the horizontal padding configuration, which is referred to as the ideal experiment setup in the rest of the paper. This setup has the following distribution for each set: 2191 patients in the training set, 547 patients in the validation set and 305 patients in the test set. The model is trained for 50 epochs with a batch size of 32. The learning rate is set to a constant value of 0.001 throughout the training. Binary cross entropy with logits loss function is used with weight of 10 for the positive class 'scar' to correct for class imbalance due to sparse existence of scar in LV segments. The results with horizontal and no padding configurations, where the model is trained on constant 50% percent of the dataset are compared in Figure 7 with the confusion matrices for the segmental level. Out of 5490 LV segments (of 305 test patients with 18 segments each), 5330 were correctly classified without the padding and 5355 were correctly classified with the horizontal padding.

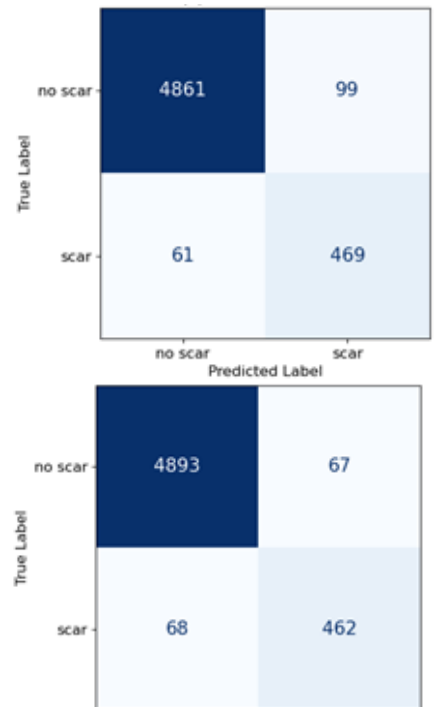

*Figure 7 Comparison of the performances of the scar detection network without padding (top) and with horizontal padding (bottom) when the dataset scale is set to 50%. There is a drop in confusion of the non-scarred segments when horizontal padding is applied.*

*Table 1 Performance evaluation of the models that were trained on the bull's eye representation with and without padding when the dataset scale is set to 50%. The scores are given for global (patient), territorial, and segmental levels.*

|  | Level | Accuracy | Balanced Accuracy | Sensitivity | Specificity |
|---|---|---|---|---|---|
| **No padding** | Patient | 0.95 | 0.95 | 0.94 | 0.97 |
|  | Left Anterior Descending (LAD) | 0.96 | 0.93 | 0.88 | 0.98 |
|  | Left Circumflex (LCx) | 0.98 | 0.98 | 0.96 | 0.99 |
|  | Right Coronary Artery (RCA) | 0.96 | 0.96 | 0.96 | 0.96 |
|  | Segments | 0.97 | 0.93 | **0.88** | 0.98 |
| **Horizontal padding** | Patient | **0.97** | **0.97** | 0.94 | **1.00** |
|  | Left Anterior Descending (LAD) | **0.98** | **0.94** | 0.88 | **1.00** |
|  | Left Circumflex (LCx) | **0.99** | 0.98 | 0.96 | **1.00** |
|  | Right Coronary Artery (RCA) | **0.99** | **0.98** | 0.96 | **0.99** |
|  | Segments | **0.98** | 0.93 | 0.87 | **0.99** |

Table 1 shows in detail the prediction scores for the same experiment setup as in Figure 7. Horizontal padding yields higher accuracy, balanced accuracy, and specificity for almost all the levels, while both options yield comparable sensitivity scores. Excellent specificity scores are achieved with the horizontal padding. Having higher specificity than sensitivity scores is expected when the class imbalance in favor of the negative class is considered. The proposed system can identify significantly higher proportion of no-scar segments, territories, and patients than it can identify the scarred ones due to having 10 times more samples with the negative class label in the training set.

Padding the LV segments in horizontal axis as shown in Figure 4C, improves all the scores except for the sensitivity score at almost all the levels. At the segmental level, sensitivity scores of 88% and 87% are achieved for no padding and horizontal padding options respectively, which are calculated directly on the segment-wise predictions of the FCN model without any post-processing. The model yields 94% sensitivity score at the global patient level for both no padding and horizontal padding, indicating that a high proportion of patients who have at least one of their LV segments scarred can be identified correctly.

The results from the selected experiment setup where the model is trained with the horizontal padding configuration of the bull's eye template on 100% of the dataset is given in Figure 8. The balanced accuracy score is given for three levels with increasing granularity from left to right. Out of 156 patients with scar, five of them were misclassified as no scar. Out of 52 patients with scar on their LAD territory, three of them were misclassified as no-scar. Out of 55 patients with scar on their LCx territory, none of them were misclassified as no-scar. Out of 49 patients with scar on their RCA territory, two of them were misclassified as no-scar. Out of 5490 LV segments (of 305 test patients with 18 segments each), 5376 were correctly classified. Out of 530 segments with scar within these 5490 segments, only 41 of them were misclassified as no-scar. In this ideal experiment setup, the accuracy and balanced accuracy scores of the model are 0.98 and 0.95 respectively. In addition, the model can correctly identify 97% of the patients with scar and 99% of the patients with no-scar. With a sensitivity score of 0.92 at the segment level and 0.97 at the patient level, the model that is trained with the ideal experiment setup outperforms the models that are trained with 50% of the dataset which are listed in Table 1.

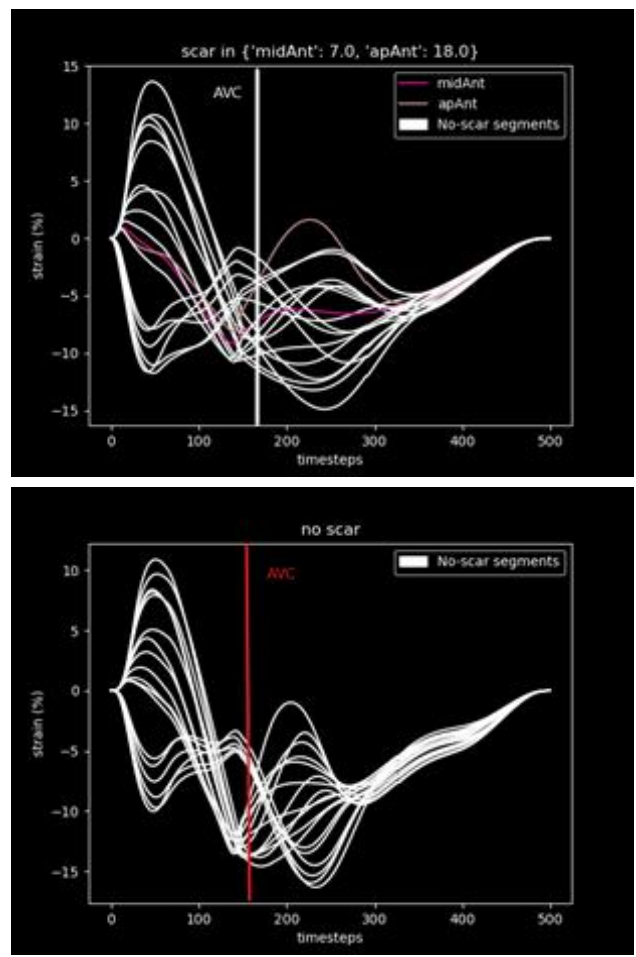

*Figure 8 The strain traces of the patients which the algorithm misclassified as no-scar (top) and scar (bottom).*

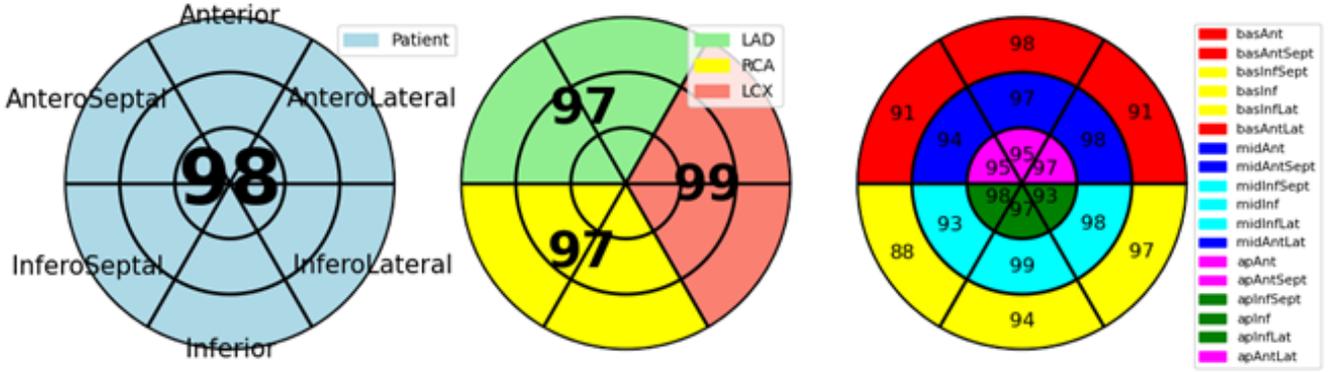

*Figure 9 The balanced accuracy scores for the model that is trained with dataset scale of 100% and the horizontal bull's eye template. The scores are reported for three levels: Global (left), territorial (middle) and segmental (right).*

Strain traces shown in Figure 3 are taken from the set of patients who are classified correctly by the proposed algorithm. It is also important to investigate the misclassified patients to understand what may have caused the false predictions. In Figure 8, the strain traces of two misclassified patients are shown with their true labels of scar existence. The patient in the top row is misclassified as no-scar patient while there are two LV segments with low volume fraction of scar as highlighted with colors in the figure. For this patient, having low volume fraction of scar may have caused the misclassification due to mild effect on the traces. The patient in the bottom row has no scar, but the algorithm predicted one of the LV segments as scarred.

**IV. DISCUSSION**

The FCN classifier trained with the unwrapped bull's eye strain resulted in excellent performance for detecting scar in LV segments of the virtual patients. Since this is a binary classification task, sensitivity is the recall of the positive 'scar' class, while specificity is the recall of the negative 'no scar' class. Thus, the sensitivity column of the score table is of importance when the priority is to correctly identify high proportion of the patients with actual scar. Having an automatic framework to identify the patients with myocardial scar with high sensitivity is expected to improve the decision-making processes in the clinics as the treatment plans for those patients need to be adapted accordingly.

While using only the echocardiography exams is not the standard procedure for LV segmental scar detection, similar [20], we investigate the possibility of identifying scar with the use of echocardiography. In addition, we try to overcome the shortcomings of manual assessment of strain traces by enabling automatic interpretation using a CNN. In [7], the authors use the GLS for scar prediction which could result in discarding the temporal patterns of the strain traces. In our study, we work with the strain traces from the whole heart cycle to preserve temporal aspect of the signals.

Both [11] and [12] overcome the issues of manual interpretation of strain traces and makes use of the temporal structure of the traces. However, the spatial representation of strain data omits the locations of LV segments in these studies. Inspired by the frequent use of bull's eye representation in clinics, we introduce a novel 2D representation of strain based on 18 segment model of the LV.

In [11], the authors reported the MI classification on the patient level, while the local damage information of coronary artery territories and LV segments are unknown. We tackle this problem by reporting the scar detection results in three different levels: global, segmental, and territorial, so that the use of our predictions can provide more detailed insight into the condition of the patient. Using an unsupervised approach, the authors in [12] eliminated the need for labeling the patient data. However, using the conventional k-means method for clustering, the additional feature extraction step for projecting the strain traces to a lower dimensional space could be eliminated, which could result in the loss of critical information. In our study, we use the raw strain traces without any processing other than the resampling operation. Although we are using a supervised learning technique, we take advantage of the virtual patient cohort which comes in pair with the automatically assigned scar labels in addition to the controlled LV strain traces.

In the presented work, which serves as a proof of concept for myocardial scar detection, a virtual patient dataset is used to train and test the model. While this provided a controlled dataset to investigate the potential of the proposed method, there are limitations of using a virtual patient dataset. Simulated strain traces do not carry the noise that can be present in the strain traces that are calculated based on estimated measurements on the ultrasound images of the patients. There are no inter-user differences present in virtual dataset; given the same input parameters, CircAdapt model is expected to behave identical in each run, when generating the patient strain data. This may suggest an easier classification task when compared to classifying strain traces of the real patients, which may then result in poor generalization performance when the learning is transferred to the real patient setup.

The cardiac cycle time markers for aortic and mitral valve closure that are used in preprocessing the strain traces are obtained from the CircAdapt model in this work. While in clinical setup, ECG is used for generating the time markers, which is subject to noise and time markers can be mislocated as a result. This, in addition to the varying quality of the images acquired during the exam may introduce additional challenges when the proposed method is going to be validated on the strain traces of the real patients.

Using one hot encoding for labeling neglects the volume fraction of scar in the LV segment, which could lead to misclassification of the segments with minor scar fractions. In the next step, using multi-class labeling to grade the scar severity should help tackling this problem. Accordingly, we would like to train the FCN for predicting different extension levels of the scar in addition to detecting their locations, by leveraging the volume fraction information.

The present work leveraged image channels in the input layer of the FCN to capture the temporal information in strain data. Alternatively, the input data could be represented by a 3D tensor, and a 3D CNN could be used to classify the strain traces for predicting the segmental scar existence, where the third dimension could be leveraged for storing temporal strain.

## V. CONCLUSION

In this work, we propose a novel method to detect myocardial scar using a multi-level approach, which corresponds to global, territorial, and segmental predictors of myocardial scar. A simple FCN model is trained and validated on a dataset of 2738 virtual patients generated by the CircAdapt model. The proposed model was able to correctly detect the absence and existence of scar in 95% of the 5490 LV segments of 305 patients in the test set, in the ideal experiment setup, where the padding configuration is set to horizontal, and the entire dataset is used. The patients who have at least one of their segments scarred were detected with 97% accuracy, while the existence of scar in coronary artery territories of the patients were detected with 94%, 100% and 96% accuracies for LAD, LCx and RCA respectively. An improved sensitivity score at the segmental level is anticipated if the model can be trained on a balanced dataset. However, realistically, the ratio of non-scarred segments is expected to be higher when compared to scarred segments in the clinics. Therefore, it is important that the proposed method can give over 0.9 sensitivity score despite the class imbalance. A multi-class labeling approach where the degree of LV scar is graded based on the volume fraction of the scar for each patient is also expected to give better performance at all levels.

In future work, this proof of concept will be validated on real patient data to investigate the applicability to clinics. As a follow-up experiment, we would like to evaluate the performance of the FCN model that is pre-trained on the virtual patient data, on the strain traces of the real patients.

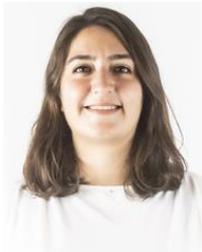

**MÜJDE AKDENİZ** received the B.Sc. degree in computer engineering from İstanbul Bilgi University, İstanbul, Turkey and the M.Sc. degree in computer engineering from Boğaziçi University, İstanbul, Turkey. She joined the cardiovascular ultrasound R&D team in GE Healthcare in April 2020. She is working as an Early Stage Researcher for the project MARCIUS funded by the European Union's Horizon 2020 research and innovation programme under the Marie-Sklodowska-Curie grant. She is studying at the PhD program of the Department of Informatics at University of Oslo, Oslo, Norway. Her research focuses on training neural networks to recognize patterns in ventricular deformation curves, that are made available via Automated Function Imaging software of GE Healthcare's ultrasound scanner.

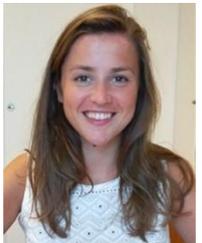

**CLAUDIA ALESSANDRA MANETTI** received the B.Sc. degree in Biomedical Engineering from the Polytechnic University of Milan (Italy) where she also got the M.Sc. degree in Information of Bioengineering. During her master, she worked as a research intern at Philips Research/Patient Care & Measurements department Eindhoven (Netherlands). In September 2020, she joined the Biomedical Engineering Department of Maastricht University under the supervision of Dr. Ir. Joost Lumens. She's working as an Early Stage Researcher for the project MARCIUS funded by the European Union's Horizon 2020 research and innovation programme under the Marie-Sklodowska-Curie grant. The goal of her project is to generate an in-silico virtual cohort of heart failure patients with the well-known CircAdapt model developed at Maastricht University.

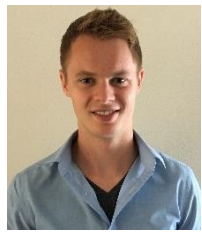

**TIJMEN KOOPSEN** received his BSc degree in Biomedical Engineering and his MSc degree in Medical Engineering at the Eindhoven University of Technology. In 2017 he started his PhD program at the Cardiovascular Research Institute Maastricht (CARIM) of Maastricht University. His research focuses on using personalized cardiac computational modeling for better identification of heart failure patients who can benefit from cardiac resynchronization therapy.

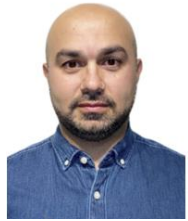

**HANI NOZARI MIRAR** was born in Mazandaran, Iran, in 1985. He received the M.Sc. degree in electronics engineering specializing on signal and image processing from the Iran University of Science and Technology, Tehran, Iran, the Ph.D. degree in photorealistic visualization of the heart the future of 3D ultrasound imaging from the Department of Informatics, University of Oslo, Norway. Since 2016, he has been working with GE HealthCare Ultrasound to do research and development on speckle tracking echocardiography.

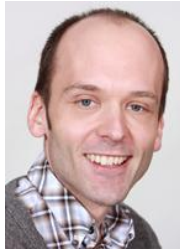

**STEN ROAR SNARE** was born in Kongsvinger, Norway, in 1981. He received the M.Sc. degree in engineering cybernetics and the Ph.D. degree in medical technology from the Norwegian University of Science and Technology (NTNU). He worked at the Department of Circulation and Medical Imaging at NTNU until fall 2011, when he was hired at GE HealthCare Ultrasound, Norway. His research interests are medical ultrasound technology, in particular image analysis for automated quantification of physiological parameters from ultrasound images.

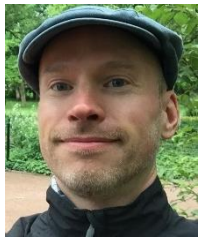

**SVEIN ARNE AASE** was born in Oslo, Norway, in 1978. He received the M.Sc. degree in computer science from the Norwegian University of Science and Technology, Trondheim, Norway, in 2003, and the Ph.D. degree from the Department of Circulation and Medical Imaging, Faculty of Medicine, Norwegian University of Science and Technology, in 2008. He spent a period as a Postdoctoral Researcher with the Department of Circulation and Medical Imaging, Norwegian University of Science and Technology, and was employed by GE HealthCare Ultrasound, Horten, Norway, in 2010. Since 2017, his main fields of interests have been deep learning applications in echocardiography.

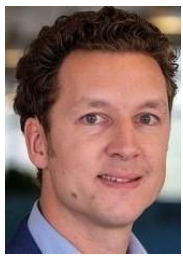

**JOOST LUMENS** obtained his BSc (Biomedical Engineering) and MSc (Medical Engineering) degrees at the Eindhoven University of Technology. In 2010, he completed a PhD (multi-scale computational modeling and simulation of cardiovascular mechanics and hemodynamics) at the Cardiovascular Research Institute Maastricht (CARIM) of Maastricht University. He continued his academic career as a post-doctoral fellow at the LIRYC Electrophysiology and Heart Modelling Institute of CHU de Bordeaux (France), where he combined computational, experimental, and clinical data to unravel the working mechanisms of cardiac pacing therapies in the dyssynchronous failing heart. Since 2012, he is working again at CARIM, where he leads the CircAdapt Research and Education team (www.circadapt.org). In 2021, he was appointed Professor of Computational Cardiology at Maastricht University Medical Center. His research team typically combines advanced computational modelling/simulation techniques with experimental and/or clinical data to gain mechanistic insight in cardiac diseases and their treatments.

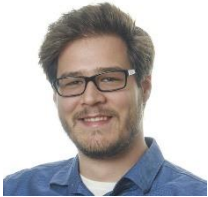

**JURICA ŠPREM** obtained his BSc (Computer Science) and MSc (Information and Communication Technology) degrees at the University of Zagreb, Faculty of Electrical Engineering and Computing. In 2019, he completed his PhD (Enhanced Cardiovascular Risk Prediction by Machine Learning) at Image Science Institute within University Medical Center Utrecht (UMC) of Utrecht University. At the same year, he joined the GE Vingmed Ultrasound department of GE Healthcare in Norway as a contracted data scientist, after which he became a tech lead for AI development within the aforementioned department and recently got deeper involved in strategic AI development within his department. His research topics not only have industrial focus as he combines other research activities and projects, (co-)supervising PhD and Master students within the AI and machine learning domain driving forward the innovations and AI initiatives within his field.

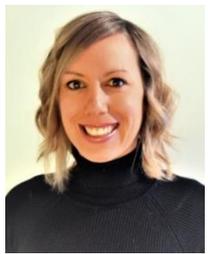

**KRISTIN SARAH MCLEOD** received a BA (double major in mathematics and psychology), BSc Honours (mathematics) and MSc (mathematics) in 2010 from Massey University in New Zealand. In 2013 she completed a PhD (artificial intelligence for medical image analysis) at the Université Côte d'Azur in France. Kristin continued her academic career at Simula Research Laboratory in Norway as a postdoctoral research fellow from 2013-2017, focusing on the development of automatic tools for supporting diagnosis and therapy planning for cardiac diseases through the use of artificial intelligence. In 2017, Kristin Sarah joined the GE Vingmed Ultrasound department of GE Healthcare in Norway as a data scientist, before transitioning to the role of Digital Manager. Through the role of Digital Manager, she has been working at the intersection of research and development, playing the role of the coordinator and principal investigator for national and EU research programs, as well as the strategic lead of the development of AI features and app ecosystem initiatives at GE Vingmed Ultrasound.